\documentclass[11pt, letterpaper]{berkeley}

\usepackage[all]{hypcap}
\usepackage[comma,numbers,sort,compress]{natbib}
\usepackage{hyperref}[citecolor=magenta]

\hypersetup{
    colorlinks = true,
    citecolor = {magenta},
}

\usepackage{microtype}
\usepackage{graphicx}
\usepackage{subfigure}
\usepackage{booktabs}
\usepackage{float}
\usepackage{xcolor}
\usepackage{amsmath}
\usepackage{amssymb}
\usepackage{mathtools}
\usepackage{amsthm}
\usepackage{mathrsfs}
\usepackage{nicefrac}
\usepackage{dsfont}
\usepackage{enumitem}
\usepackage{float}

\usepackage{xspace}
\usepackage[capitalize,noabbrev]{cleveref}
\usepackage{subcaption}
\usepackage{wrapfig}
\usepackage{lipsum}
\usepackage{listings}

\usepackage{amsmath}
\usepackage{amssymb}
\usepackage{mathtools}
\usepackage{amsthm}
\usepackage{bbm}
\usepackage{tabularx}
\usepackage{multirow}
\usepackage[export]{adjustbox}
\usepackage{colortbl}
\usepackage{soul}

\usepackage{setspace}

\usepackage{hyperref}
\usepackage{url}
\usepackage{microtype}
\usepackage{amsmath}
\usepackage{amssymb}
\usepackage{bbm}
\usepackage{xcolor}
\usepackage[ruled]{algorithm2e}
\usepackage{graphicx}
\usepackage{wrapfig}
\usepackage{booktabs}
\usepackage{pifont} 
\usepackage[dvipsnames]{xcolor}
\usepackage{color}
\sethlcolor{olive} 
\definecolor{deepblue}{rgb}{0,0,0.5}
\definecolor{deepred}{rgb}{0.6,0,0}
\definecolor{deepgreen}{rgb}{0,0.5,0}
\definecolor{boost_correct_to_correct}{HTML}{66C2A5}
\definecolor{default_correct_to_correct}{HTML}{fc8d62}
\definecolor{dup_correct_to_correct}{HTML}{8da0cb}
\definecolor{new_correct_to_correct}{HTML}{e78ac3}

\newcommand\pythonstyle{\lstset{
basicstyle=\ttfamily\footnotesize,
language=Python,
morekeywords={self, clip, exp, mse_loss, uniform_sample, concatenate, logsumexp},              
keywordstyle=\color{deepblue},
emph={MyClass,__init__},          
emphstyle=\color{deepred},   
stringstyle=\color{deepgreen},
frame=single,                       
showstringspaces=false
}}

\lstnewenvironment{python}[1][]
{
\pythonstyle
\lstset{#1}
}
{}

\newcommand\pythoninline[1]{{\pythonstyle\lstinline!#1!}}

\DeclareRobustCommand{\hlg}[1]{{\sethlcolor{YellowGreen}\hl{#1}}}

\definecolor{blanchedalmond}{rgb}{1.0, 0.92, 0.8}
\definecolor{carmine}{rgb}{0.59, 0.0, 0.09}
\definecolor{lightblue}{rgb}{0.22,0.45,0.70}

\renewcommand{\mathbf}{\boldsymbol}

\makeatletter
\def\Ddots{\mathinner{\mkern1mu\raise\p@
\vbox{\kern7\p@\hbox{.}}\mkern2mu
\raise4\p@\hbox{.}\mkern2mu\raise7\p@\hbox{.}\mkern1mu}}
\makeatother

\newcommand{\ours}{\texttt{SIKeD}}
\newcommand{\oursfull}{\texttt{SIKeD}: \textbf{S}elf-guided \textbf{I}terative \textbf{K}nowledge \textbf{D}istillation}

\newcommand{\question}{q}
\newcommand{\answer}{a}
\newcommand{\rationale}{r}
\newcommand{\reasoningstrategy}{s}
\newcommand{\reasoningstrategies}{S}

\numberwithin{equation}{section}

\definecolor{amaranth}{rgb}{0.9, 0.17, 0.31}
\definecolor{antiquebrass}{rgb}{0.8, 0.58, 0.46}
\definecolor{antiquefuchsia}{rgb}{0.57, 0.36, 0.51}
\definecolor{chromeyellow}{rgb}{0.31, 0.47, 0.26}

\usepackage[dvipsnames]{xcolor}
\definecolor{maj5}{HTML}{2b8cbe}
\definecolor{maj5Imp}{HTML}{084081}

\definecolor{seq5wo}{HTML}{d95f0e}
\definecolor{seq5woImp}{HTML}{662506}

\definecolor{seq5w}{HTML}{6a51a3}
\definecolor{seq5wImp}{HTML}{3f007d}

\definecolor{selfwo}{HTML}{d95f0e}
\definecolor{selfwoImp}{HTML}{662506}

\definecolor{selfw}{HTML}{6a51a3}
\definecolor{selfwImp}{HTML}{3f007d}

\definecolor{glorewo}{HTML}{d95f0e}
\definecolor{glorewoImp}{HTML}{662506}

\definecolor{glorew}{HTML}{6a51a3}
\definecolor{glorewImp}{HTML}{3f007d}

\definecolor{vstar}{HTML}{d95f0e}
\definecolor{vstarImp}{HTML}{662506}

\makeatletter
\def\mathcolor#1#{\@mathcolor{#1}}
\def\@mathcolor#1#2#3{%
  \protect\leavevmode
  \begingroup
    \color#1{#2}#3%
  \endgroup
}
\makeatother

\usepackage[textsize=tiny]{todonotes}
\usepackage{amssymb}

\usepackage[skins,theorems]{tcolorbox}
\Crefformat{equation}{#2Eq.\;(#1)#3}

\Crefformat{figure}{#2Figure #1#3}
\Crefformat{assumption}{#2Assumption #1#3}
\Crefname{assumption}{Assumption}{Assumptions}

\usepackage{crossreftools}
\usepackage[suppress]{color-edits}

\usepackage{dsfont}
\usepackage{nicefrac}

\newtcolorbox{analysisbox}[1][]{
    enhanced jigsaw,
    colback=white,
    colframe=blue!75!black,
    fonttitle=\bfseries,
    boxsep=5pt,
    left=5pt,
    right=5pt,
    top=5pt,
    bottom=5pt,
    title=#1,
}

\usepackage{pifont}
\usepackage{soul}
\definecolor{highlightmistake}{RGB}{255, 179, 179}
\definecolor{highlightcorrect}{RGB}{179, 255, 179}

\title{\texttt{SIKeD}: \textbf{S}elf-guided \textbf{I}terative \textbf{K}nowledg\textbf{e} \textbf{D}istillation for mathematical reasoning}

\author[1]{Shivam Adarsh*}
\author[2]{Kumar Shridhar*}
\author[3]{Caglar Gulcehre}
\author[4]{Nicholas Monath}
\author[2]{Mrinmaya Sachan}

\affil[1]{University of Zurich}
\affil[2]{ETH Zurich}
\affil[3]{EPFL}
\affil[4]{Google Deepmind}

\correspondingauthor{\texttt{(*)} equal contribution. Correspondence at: \{shivam.adarsh@uzh.ch ; shkumar@ethz.ch\}  }

\begin{abstract}
Large Language Models (LLMs) can transfer their reasoning skills to smaller models by teaching them to generate the intermediate reasoning process required to solve multistep reasoning tasks.  While LLMs can accurately solve reasoning tasks through a variety of strategies, even without fine-tuning, smaller models are not expressive enough to fit the LLMs distribution on all strategies when distilled and tend to prioritize one strategy over the others.
This reliance on one strategy poses a challenge for smaller models when attempting to solve reasoning tasks that may be difficult with their preferred strategy.
To address this, we propose a distillation method \oursfull \footnote{\ours{} is pronounced as ``psyched''}, where the LLM teaches the smaller model to approach a task using different strategies and the smaller model uses its self-generated on-policy outputs to choose the most suitable strategy for the given task. The training continues in a \emph{self-guided} iterative manner, where for each training iteration, a decision is made on how to combine the LLM data with the self-generated outputs. Unlike traditional distillation methods, \ours\ allows the smaller model to learn \emph{which} strategy is suitable for a given task while continuously learning to solve a task using different strategies.
Our experiments on various mathematical reasoning datasets show that \ours\ significantly outperforms traditional distillation techniques across smaller models of different sizes.

\vspace{0.25em}
\hspace{0.7em}\includegraphics[width=1.0em,height=1.0em]{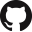}\hspace{.5em}\parbox{\dimexpr\linewidth-2\fboxsep-2\fboxrule}{\url{https://github.com/kumar-shridhar/SIKeD}}
\vspace{-.5em}
\end{abstract}

\begin{document}
\maketitle

\section{Introduction}

\begin{wrapfigure}{r}{0.48\textwidth}  
    \centering
    \vspace{-1.1cm}
    \includegraphics[width=0.48\textwidth]{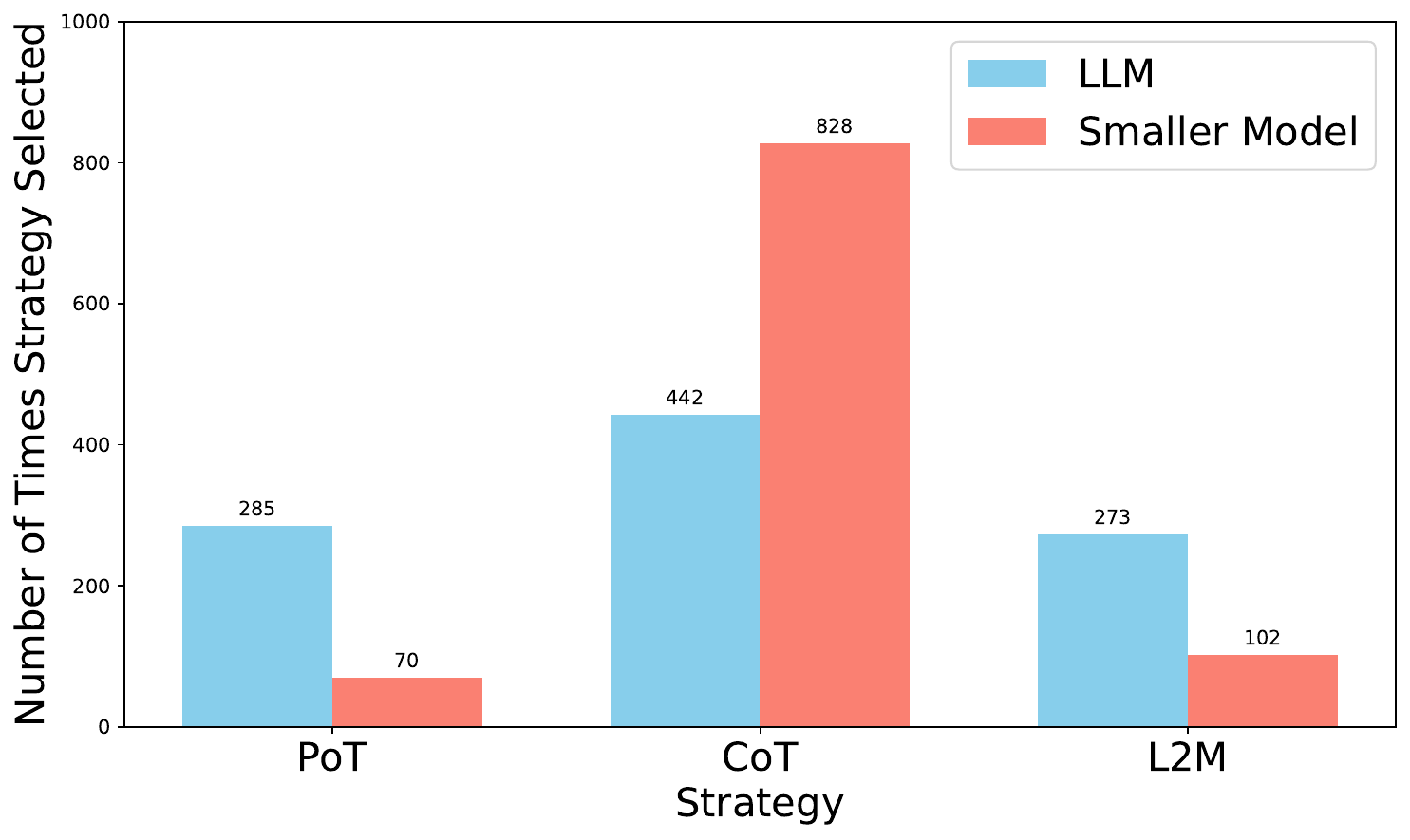}
    \vspace{-10pt}
    \caption{\textbf{Histogram of strategy choices for the LLM and the smaller model.} \textcolor{Cerulean}{LLM} tends to select several reasoning strategies, but \textcolor{RedOrange}{the smaller model} is biased towards one strategy. The comparison was done on 1000 data points randomly sampled from the GSM8K train set.}
    \label{fig:main-plot}
    \vspace{-0.3cm}
\end{wrapfigure}

Large language models (LLMs), with tens to hundreds of billions of parameters, generally outperform smaller models (with billions of parameters or fewer) in a variety of reasoning tasks \cite{llama, gpt4}. One notable strength of large models is their ability to reason and perform multistep reasoning tasks, often considered an important aspect of intelligence \cite{reasoning-as-intelligencve}. However, the significant size and computational demands of these large models present several challenges. For example, LLaMA3 models \cite{llama} are trained using clusters of 24,000 GPUs, limiting their accessibility to many researchers and practitioners. 

To bridge this gap, a key approach involves teaching smaller models to replicate the knowledge of a larger model, often referred to as \emph{knowledge distillation} \cite{hinton2015distilling}. Typically, smaller models can be taught to replicate the multistep reasoning capabilities of larger models by incorporating a set of intermediate sequences \citep{kim-rush-2016-sequence, shridhar-etal-2023-distilling}.
However, these intermediate steps can be derived from a number of different strategies, such as Chain of Thought (CoT) \cite{cot}, Subquestion Decomposition \cite{shridhar-automatic, l2m}, and Program of Thoughts (PoT) \cite{pot}, among others.
A viable solution is to distill these reasoning capabilities into smaller models either by distilling individual strategies \citep{magister-etal-2023-teaching, shridhar-etal-2023-distilling, hsieh-etal-2023-distilling} or by incorporating multiple strategies simultaneously \citep{chenglin2023mixed, zhu2024improvingmath}. 
Although smaller models have demonstrated impressive performance when distilled with a single strategy, they often struggle to master multiple strategies equally well. An example is presented in \autoref{fig:main-plot} where a larger model can use multiple strategies to generate data but upon distilling, a smaller model tends to favor one over the others. This is because reasoning through a variety of strategies tends to emerge as a result of scaling language models, making it difficult for smaller models to replicate this behavior \cite{lyu2024calibrating}. 

On the other hand, learning to solve a task using multiple strategies can help smaller models overcome the limitations of relying on a single approach. However, a key challenge arises when, despite being trained on a fixed dataset containing various strategies, a distribution mismatch occurs between the data generated by the LLM and the outputs produced by the smaller model during inference. This mismatch can hinder the ability of the smaller model to generalize across different reasoning strategies. This issue, often discussed in imitation learning \citep{pomerleau1991efficient, pmlr-v9-ross10a}, results in the student model consistently choosing one strategy, even when a different approach would be more appropriate. As a result, the student generates outputs with strategy choices that are highly unlikely to match those produced by the teacher.

To address this challenge, we introduce our distillation methodology, \oursfull. The process begins with the LLM teaching the smaller model to approach tasks using a variety of reasoning strategies, providing a strong foundation for the smaller model to understand different problem-solving approaches. However, due to inherent biases and its limited capacity, the smaller model may still struggle to match the LLM's distribution of strategy choices effectively.
To resolve this, we take inspiration from constructivist learning theory \cite{narayan2013constructivism}, where the learner builds knowledge during the ``assimilation phase'' and refines their understanding during the ``accommodation phase'' to incorporate new insights.
We propose generating outputs using the smaller model in an on-policy setup and selecting the best strategies for the task. By mixing the LLM-generated data with self-generated outputs, we leverage the strengths of both datasets. We iteratively fine-tune the smaller model allowing it to recognize strategies that it learned from the LLM but did not initially apply. With this approach, we align the smaller model with its own learned knowledge rather than forcing its distribution to mirror that of the LLM’s.

Our proposed method extends beyond traditional one-step distillation, as each iteration of \ours\ leads to an updated policy that better grasps new information. We repeat multiple iterations of \ours\ based on the accuracy-cost tradeoff (does the improvement justify the cost of another iteration), allowing for continuous refinement and improvement of the model's reasoning capabilities.
We demonstrate the effectiveness of \ours\ on several mathematical reasoning tasks using models with fewer than 7 billion parameters.\footnote{We acknowledge that ``smaller model'' is a relative term, and we consider models with fewer than 7 billion parameters to be smaller models.} On four mathematical datasets—GSM8K \cite{cobbe2021training}, SVAMP \cite{patel-etal-2021-nlp}, ASDiv \cite{miao-etal-2020-diverse}, and MultiArith \cite{roy2016solving}—our approach achieves improvements of up to +5 points over traditional distillation strategies. Additionally, we show that multiple rounds of \ours\ allow the model to select the appropriate strategy for a given problem, while traditional distillation using LLM's data tends to leave it biased.

\section{Preliminaries: LLM based Distillation}

\paragraph{Problem Setup}
We consider the standard setup of LLM-based distillation (also referred to as \emph{knowledge distillation}),  where the data is sampled from the larger model (LLM) with intermediate reasoning and the smaller model is fine-tuned (distilled) over it \cite{shridhar-etal-2023-distilling, magister-etal-2023-teaching}. Two auto-regressive sequence models are involved in the process: a larger model or the LLM denoted as $p_L$ and a smaller model to be distilled as $p_{\texttt{sm}}^\theta$ (with learnable parameters $\theta$). In this work, we consider a reasoning dataset $\mathcal{D}$ consisting of a question $\question_i$ and a numerical answer $\answer_i$ pairs for $n$ data points, i.e. $i \in \{1, \ldots, n\}$. Since our work focuses on improving reasoning in smaller models by teaching them to solve a variety of reasoning strategies, we consider three reasoning strategies in this work: Chain-of-Thought (\texttt{CoT}), Least-to-Most (\texttt{L2M}), and Program-of-Thought (\texttt{PoT}). 
For a specific \emph{reasoning strategy}, denoted as $\reasoningstrategy \in \reasoningstrategies$, we generate the reasoning chain or \emph{rationale}, denoted as $\rationale_i$ leading to the final answer as: $\rationale_i \sim p_{L}(. \mid \texttt{pr}_{\reasoningstrategy} , \question_i)$,
where, $\texttt{pr}_{\reasoningstrategy}$ represents the  strategy-specific prompt, and  $\reasoningstrategy \in \{\texttt{CoT}, \texttt{L2M}, \texttt{PoT}\}$. Prompts used are provided in \autoref{app:prompts}.

\subsection{LLM based Distillation}
We add the generated rationales to the the dataset $\mathcal{D}$ to create an initial training dataset  $\mathcal{D}_{\texttt{LLM}}$ consisting of a quadruple of $\{\question_i, \answer_i, \reasoningstrategy, \rationale_i\}$ for each data point. 
We perform a data filtering by extracting the final answer $\hat{\answer}_i$ from the generated rationale $\rationale_i$ and comparing it with the ground truth answer $\answer_i$. We discard all samples that do not match, i.e., we keep samples where $\hat{\answer}_i = \answer_i$. This filtering process eliminates incorrect rationales, ensuring that only high-quality data is used for distilling the smaller models. 

We start the distillation process by training the smaller model with the created dataset $\mathcal{D}_{\texttt{LLM}}$. The question $\question_i$ is provided as input, and the smaller model $p_{\texttt{sm}}^\theta$ (with learnable parameters $\theta$) is first instructed to generate the strategy $\reasoningstrategy$, followed by the rationale $\rationale_i$ that leads to the final answer $\answer_i$. The loss $\mathcal{L}_{\texttt{L}}(\theta)$ is defined as:

\begin{equation}
    \label{eq:nll}
    \mathcal{L}_{\texttt{L}}(\theta) = - \mathbb{E}_{(\question_i, \reasoningstrategy, \rationale_i) \sim \mathcal{D}_{\texttt{LLM}}} \left[ \log\ p_{\texttt{sm}}^\theta\ (\reasoningstrategy \mid \question_i, I) + \sum_{t=1}^{M} \log p_{\texttt{sm}}^\theta\ (\rationale_{i,t} \mid \rationale_{i,< t}, \reasoningstrategy, \question_i, I) \right],
\end{equation}

where $M$ represents the number of tokens decoded over time $t$ in an autoregressive manner, and $I$ is the instruction used during fine-tuning. Note that this is analogous to traditional knowledge distillation from LLMs except that we make a strategy choice before generating rationales.

\paragraph{Limitations}
Training solely on LLM-generated data $\mathcal{D}_{\texttt{LLM}}$ can lead to a distribution mismatch between the training data and the smaller model's own output distribution. Specifically, the larger model due to its larger capacity, may produce correct reasoning across multiple strategies that the smaller model can find difficult to replicate directly \cite{agarwal2024onpolicy}. A comparison of the strategy selected by the LLM and the smaller model on 1K samples is presented in \autoref{fig:main-plot}. The smaller model performs poorly when generating outputs on its own, as the training data distribution $P_{\text{train}}(x)$ is different from the model's output distribution $P_{\texttt{sm}}^{\theta}(x)$  as $P_{\text{train}}^{(1)}(x) = P_{\text{LLM}}(x),$ where $x$ represents the samples $(\question_i, \reasoningstrategy, \rationale_i)$, and $P_{\text{LLM}}(x)$ is the distribution of data generated by the LLM $p_L$.

\paragraph{Proposed Solution}
To mitigate the distributional shift in strategy choice between the LLM and the smaller model, we propose to incorporate the smaller model's own correct outputs into the training data. This \emph{self-guided} training with data mixing aligns the training data distribution more closely with the smaller model's output distribution, making learning more effective. A visualization of the data mixing approach is presented in \autoref{fig:main-fig} that demonstrates that data mixing reduces the distribution shift, bringing the LLM and the smaller model's output distribution closer. This allows the smaller model to choose the right strategy for a given task, much like the LLM.
\section{\oursfull}

We propose \ours, an \emph{iterative} training where smaller models can take advantage of their own generations to refine their strategy choices for a given task. In a nutshell, we generate data from the smaller model, filter out the correct samples based on whether the generated solutions are correct, and mix this data with the LLM-generated data to adjust its strategy preferences. The smaller distilled model is used to iteratively generate data in an on-policy setting where it updates itself by leveraging both the LLM data and its own generations. This iterative process  allows the smaller model to improve its reasoning abilities and strategy selection over time by leveraging the LLM's knowledge and its own prior learning. The following paragraphs discuss the steps involved in our proposed iterative distillation methodology and the training objective.

\paragraph{Data generation}
For each question $\question_i$ and its associated reasoning strategy $\reasoningstrategy$, we first generate $K$ rationales using the current smaller model $p_{\texttt{sm}}^\theta$ as: $
\rationale_i^{(k)} \sim p_{\texttt{sm}}^\theta (\cdot \mid \reasoningstrategy, \question_i, I), \quad \text{for } k = 1, \ldots, K.
$
Note that we generate multiple samples $K$ 
as the likelihood of a correct answer being present in one of the rationales increases significantly with additional generations for smaller models \citep{jain2023first, wang2023selfconsistency}.

\paragraph{Data Filtering}
Next, we extract the predicted answer $\hat{\answer}_i^{(k)}$ from each rationale $\rationale_i^{(k)}$ and compare it with the ground truth $\answer_i$. We collect the correct samples, where $\hat{\answer}_i^{(k)} = \answer_i$, into a new dataset $
\mathcal{D}_{\texttt{self}} = \left\{ (\question_i, \reasoningstrategy, \rationale_i^{(k)}) \ \middle| \ \hat{\answer}_i^{(k)} = \answer_i \right\}.
$

\begin{figure}[t]
    \centering
    \includegraphics[width=1.0\linewidth]{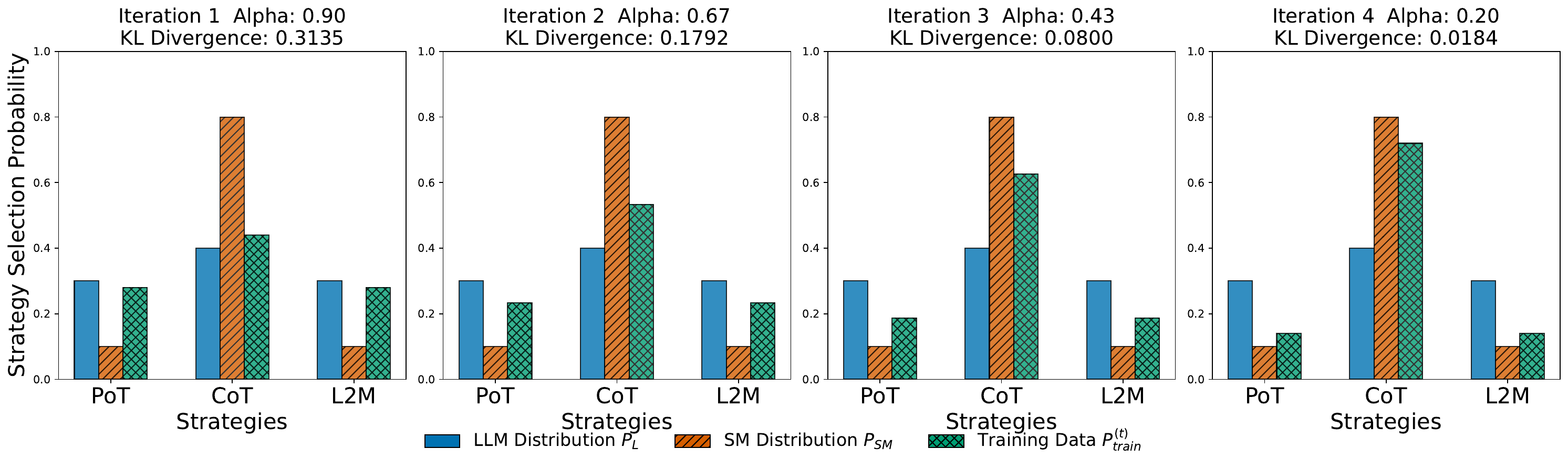}
    \caption{\textbf{Alignment of the smaller model's strategy distribution with the LLM over iterations}.  Each subplot represents an iteration in the training process, showing the probability distributions over reasoning strategies: PoT, L2M, and CoT. The \textcolor{MidnightBlue}{blue bars} depict the LLM's distribution $P_{L}$, while the \textcolor{orange}{orange bars} represent the smaller model's distribution $P_{SM}$, which is biased towards \texttt{CoT}. The \textcolor{ForestGreen}{green bars} show the training data distribution $P_{\text{train}}^{(t)}$, a mixture of $P_{L}$ and $P_{SM}$ weighted by the mixing rate $\alpha$. As $\alpha$ decreases over iterations (from 0.90 to 0.20), $P_{\text{train}}^{(t)}$ shifts from being similar to the LLM's distribution towards the smaller model's distribution. The KL divergence between the training data and the smaller model distributions decreases accordingly, indicating increased similarity. 
}
    \label{fig:main-fig}
\end{figure}

\paragraph{Data mixing}
We combine the LLM-generated dataset $\mathcal{D}_{\texttt{LLM}}$ with the self-generated dataset $\mathcal{D}_{\texttt{self}}$ to form the mixed dataset $\mathcal{D}_{\texttt{mix}} = \mathcal{D}_{\texttt{LLM}} \cup \mathcal{D}_{\texttt{self}}$.

Note that, we do not always use all the data from LLM in $\mathcal{D}_{\texttt{mix}}$, and study two variations: 
\emph{All} when all LLM data is used in $\mathcal{D}_{\texttt{mix}}$, and \emph{Adaptive} when only queries that have no correct generations in $\mathcal{D}_{\texttt{self}}$ are taken from  $\mathcal{D}_{\texttt{LLM}}$. \emph{Adaptive} uses less generated data from the LLM, resulting in more computationally efficient training.  

The corresponding training data distribution changes to a mixture of the LLM data distribution and the model's own output distribution:

\[
P_{\text{train}}^{(2)}(x) = \alpha P_{\text{LLM}}(x) + (1 - \alpha) P_{\texttt{sm}}^{\theta}(x),
\]

where $\alpha = \frac{|\mathcal{D}_{\texttt{LLM}}|}{|\mathcal{D}_{\texttt{LLM}}| + |\mathcal{D}_{\texttt{self}}|}$ serves as a normalized mixing rate between the two datasets.

\paragraph{Training objective}
By including $\mathcal{D}_{\texttt{self}}$ in the training data, we reduce the divergence between $P_{\text{train}}^{(2)}(x)$ and the model's own output distribution $P_{\texttt{sm}}^{\theta}(x)$, thus minimizing the distribution shift and improving training effectiveness of choosing the right strategy for a given task. 

We continue training the smaller model on $\mathcal{D}_{\texttt{mix}}$ using the following loss function:

\begin{equation}
    \label{eq:nll_mix}
    \mathcal{L}_{\texttt{mix}}(\theta) = - \mathbb{E}_{(\question_i, \reasoningstrategy, \rationale_i) \sim \mathcal{D}_{\texttt{mix}}} \left[ \log p_{\texttt{sm}}^{\theta}\ (\reasoningstrategy \mid \question_i, I) + \sum_{t=1}^{M} \log p_{\texttt{sm}}^{\theta}\ (\rationale_{i,t} \mid \rationale_{i,< t}, \reasoningstrategy, \question_i, I) \right]
\end{equation}

The expected loss over the training data is:
\[
\mathcal{L}_{\texttt{mix}}(\theta) = - \mathbb{E}_{x \sim P_{\text{train}}^{(2)}(x)} \left[ \log p_{\texttt{sm}}^{\theta}(x) \right]
\]

where $x = (\question_i, \reasoningstrategy, \rationale_i)$, and $p_{\texttt{sm}}^{\theta}(x)$ denotes the probability assigned by the model to the sample $x$.

\paragraph{Analogous to minimizing the KL divergence}
Mixing the data is analogous to minimizing the Kullback-Leibler (KL) divergence \cite{kullback1951information} between the training data distribution $P_{\text{train}}^{(2)}(x)$ and the model's output distribution $P_{\texttt{sm}}^{\theta}(x)$:
\[
D_{\text{KL}}(P_{\text{train}}^{(2)}(x) \parallel P_{\texttt{sm}}^{\theta}(x)) = \sum_x P_{\text{train}}^{(2)}(x) \log \frac{P_{\text{train}}^{(2)}(x)}{P_{\texttt{sm}}^{\theta}(x)}.
\]

As we include more self-generated data, $(1 - \alpha)$ increases, and $P_{\text{train}}^{(2)}(x)$ becomes closer to $P_{\texttt{sm}}^{\theta}(x)$. This reduces the KL divergence and aligns the training data distribution with the model output distribution, leading to more effective learning. \autoref{fig:main-fig} demonstrates that as the value of $\alpha$ decreases, the distribution of the training data strategy choices aligns with the smaller model, with a reduction in their KL value over iterations. This allows the smaller model to better capture the strategy distribution of the larger model.

\subsection{Iterative Self-Training of \ours}

We repeat the data generation, filtering, mixing, and training steps iteratively. In each iteration $t$, the smaller model potentially generates new correct rationales that are added to the training data. The training data distribution at iteration $t$ becomes:

\[
P_{\text{train}}^{(t)}(x) = \alpha^{(t)} P_{\text{LLM}}(x) + (1 - \alpha^{(t)}) P_{\texttt{sm}}^{\theta^{(t-1)}}(x),
\]

where $\theta^{(t-1)}$ are the model parameters from the previous iteration, and $\alpha^{(t)}$ is updated based on the sizes of $\mathcal{D}_{\texttt{LLM}}$ and $\mathcal{D}_{\texttt{self}}^{(t)}$ at iteration $t$. Note that the generated samples from the smaller model automatically govern the value of $\alpha^{(t)}$ based on the size of $\mathcal{D}_{\texttt{self}}^{(t)}$.

This iterative process continues until the model's performance converges or a predefined stopping criterion is met. Over multiple iterations, the model's own output distribution $P_{\texttt{sm}}^{\theta^{(t)}}(x)$ gradually improves, and the training data distribution becomes increasingly aligned with it. We present an end-to-end training methodology in Algorithm \ref{algo:self_training}.

\RestyleAlgo{ruled}

\begin{algorithm}[t]
\fontsize{9}{9.5}\selectfont
\KwIn{
$\mathcal{D}$: Reasoning dataset with questions $\{\question_i\}_{i=1}^N$ and answers $\{\answer_i\}_{i=1}^N$, 
$\mathcal{D}_{\texttt{LLM}}$: Reasoning dataset generated using the LLM with questions $\{\question_i\}$,  answers $\{\answer_i\}$, strategy $\{\reasoningstrategy\}$, rationales $\{\rationale_i\}$, 
$\reasoningstrategies$: Set of reasoning strategies, 
$I$ : instruction, 
$p_{L}$: LLM for rationale generation , 
$p_{\texttt{sm}}^{\theta^{(0)}}$: Smaller model with initial parameters $\theta^{(0)}$ ,
$K$: Number of samples per question and strategy,
$T$: Maximum number of iterations,
\textbf{Variation}: \emph{All} or \emph{Adaptive},

}

\BlankLine

\textcolor{blue}{\texttt{// LLM-Based Distillation}} \\

Train $p_{\texttt{sm}}^{\theta^{(0)}}$ on $\mathcal{D}_{\texttt{LLM}}$ by minimizing $\mathcal{L}_{\texttt{L}}(\theta^{(0)})$ (~\eqref{eq:nll}) \\

\BlankLine

\textcolor{red}{\texttt{// \oursfull}} \\

\For{iteration $t = 1$ to $T$}{
    \textbf{Initialize} dataset $\mathcal{D}_{\texttt{self}}^{(t)} \leftarrow \emptyset$ \\
    \For{each question $\question_i \in \mathcal{D}$}{
        \For{each strategy $\reasoningstrategy \in \reasoningstrategies$}{
            \For{$k = 1$ to $K$}{
                Generate rationale $\rationale_i^{(k)}$ using $p_{\texttt{sm}}^{\theta^{(t-1)}}$: $\rationale_i^{(k)} \sim p_{\texttt{sm}}^{\theta^{(t-1)}}(\cdot \mid \reasoningstrategy, \question_i, I)$ \\
                Extract answer $\hat{\answer}_i^{(k)}$ from $\rationale_i^{(k)}$ \\
                \If{$\hat{\answer}_i^{(k)} = \answer_i$}{
                    Add $(\question_i, \reasoningstrategy, \rationale_i^{(k)})$ to $\mathcal{D}_{\texttt{self}}^{(t)}$
                }
            }
        }
    }
    \eIf{\textbf{Variation} is \emph{All}}{
        Combine datasets: $\mathcal{D}_{\texttt{mix}}^{(t)} = \mathcal{D}_{\texttt{LLM}} \cup \mathcal{D}_{\texttt{self}}^{(t)}$ \\
    }{
        Identify questions with no correct self-generated rationales: $\mathcal{I} = \{ i \mid \text{no correct } \rationale_i^{(k)} \text{ in } \mathcal{D}_{\texttt{self}}^{(t)} \}$ \\
        Include corresponding LLM data: $\mathcal{D}_{\texttt{LLM}}^{(t)} = \{ (\question_i, \reasoningstrategy, \rationale_i) \in \mathcal{D}_{\texttt{LLM}} \mid i \in \mathcal{I} \}$ \\
        Combine datasets: $\mathcal{D}_{\texttt{mix}}^{(t)} = \mathcal{D}_{\texttt{LLM}}^{(t)} \cup \mathcal{D}_{\texttt{self}}^{(t)}$ \\
    }
    Update $\alpha^{(t)} = \frac{|\mathcal{D}_{\texttt{LLM}}^{(t)}|}{|\mathcal{D}_{\texttt{LLM}}^{(t)}| + |\mathcal{D}_{\texttt{self}}^{(t)}|}$ \\
    Retrain $p_{\texttt{sm}}^{\theta^{(t)}}$ on $\mathcal{D}_{\texttt{mix}}^{(t)}$ by minimizing $\mathcal{L}_{\texttt{mix}}^{(t)}(\theta^{(t)})$ (~\eqref{eq:nll_mix}) \\
}

\KwOut{Updated smaller model $p_{\texttt{sm}}^{\theta^{(T)}}$}
\caption{\textbf{\oursfull}}
\label{algo:self_training}
\end{algorithm}

\section{Experimental Details}

\paragraph{Dataset}
Our work demonstrates the effectiveness of selecting an appropriate strategy for a given task. We consider multi-step mathematical reasoning datasets in our work, as various strategies can solve the task fairly well. 
We trained \ours{} on the GSM8K training set \cite{cobbe2021training}, which includes 7,473 samples, and tested it on the corresponding test set of 1,319 samples. To assess the domain transferability of our distillation method, we also evaluated it on three additional mathematical datasets: SVAMP \cite{patel-etal-2021-nlp} with 1,000 samples, ASDiv \cite{miao-etal-2020-diverse} with 2,300 test samples, and MultiArith \cite{roy2016solving} with 180 samples. As the GSM8K training set was used to train the smaller model, we classify it as an \emph{in-distribution} dataset. In contrast, no training data from SVAMP, ASDiv, or MultiArith was used, as they were exclusively employed for testing purposes and thus considered \emph{out-of-distribution}.  

\paragraph{Implementation Details}
We used the Llama3 70B model \cite{llama3} as the large language model (LLM) to generate the rationales. We performed distillation on different smaller models ranging from 0.5B to 7B parameters, including Qwen2 0.5B \cite{qwen}, Qwen2 1.5B \cite{qwen}, SmolLM 1.7B \cite{huggingface2023smollm}, Gemma 2B \cite{gemma}, and Gemma 7B \cite{gemma}. All smaller models were fine-tuned using LoRA \cite{hu2021lora} with a rank of 16, and alpha of 32. We used a learning rate of 3e-4 for Qwen models with a cyclic scheduler, while we set 2e-4 as the learning rate for other models and used a linear scheduler. We train all models for 3 epochs. We implemented all our experiments using the Unsloth FastLanguageModel \cite{unslothai_unsloth} and used the VLLM library  \cite{vllm} for inference. We set the temperature $t=0$ for data generation from the LLM while $t=0.7$ was used for generating samples from the smaller model at each iteration. We set the number of generated samples or $K$ to 10. We report Top-1 accuracy (maj@1).

Our proposed approach was compared against a set of single-strategy distillation methods. In this work, we employed three reasoning strategies: Chain-of-Thought (CoT) \cite{cot}, Program-of-Thought (PoT) \cite{pot}, and Least-to-Most (L2M) \cite{l2m}.
\section{Results and Discussion}

\begin{table*}[t!]
\centering %
\small
\caption{Top-1 (maj@1) accuracy comparison across different sized models (Gemma 2B and 7B, SmolLM 1.7B, Qwen 0.5B and 1.5B) on four mathematical datasets: GSM8K, ASDiv, MultiArith, and SVAMP. ``Combined'' refers to the scenario where data from all three reasoning strategies are merged and then used for distillation. We report the two variants of \ours: ``Adaptive'' and ``All'', and the best performance is \hlg{highlighted}. \\}
{ %
\begin{tabular}{l| l | c | c | c | c}
\hline
& & \multicolumn{4}{c}{\bf Dataset} \\
\cline{3-6} 
\multicolumn{1}{c|}{\bf Model} & \multicolumn{1}{c|}{\bf Method} & \textbf{{In Distribution}} & \multicolumn{3}{c}{\textbf{Out of Distribution}} \\
\cline{4-6} 
& & GSM8K  & ASDiv & MultiArith & SVAMP\\
\hline
\hline
 & CoT & 67.40 & 68.76 & 98.33 & 66.80 \\
 & L2M & 69.29 & 64.69 & 96.11 & 64.80 \\
 & PoT & \underline{71.34} & 67.85 & 98.89 & \underline{75.00} \\
Gemma 7B & Combined & 70.74 & \underline{69.11} & \underline{99.44} & 69.40 \\
 & \ours\ & & & & \\
 & \quad Adaptive & \textbf{73.84} \hlg{{\scalebox{0.7}{($\uparrow$ +2.5)}}} & 70.59 \scalebox{0.7}{($\uparrow$ +1.5)} & 99.44 \scalebox{0.7}{(-)} & 72.90 \scalebox{0.7}{($\downarrow$ -2.1)} \\
 & \quad All & 71.42 \scalebox{0.7}{($\uparrow$ +0.1)} & \textbf{70.76}  \hlg{{\scalebox{0.7}{($\uparrow$ +1.6)}}} & \textbf{100.0}  \hlg{{\scalebox{0.7}{($\uparrow$ +0.6)}}} & \textbf{76.40}  \hlg{{\scalebox{0.7}{($\uparrow$ +1.4)}}} \\
\hline
\hline
 & CoT & 36.54 & 54.01 & 87.22 & 41.90 \\
 & L2M & 36.92 & 43.47 & 81.67 & 31.60 \\
 & PoT & 44.05 & \underline{58.13} & \underline{90.56} & \underline{56.80} \\
Gemma 2B & Combined & \underline{44.05} & 57.96 & 84.44 & 56.20 \\
 & \ours\ & & & & \\
 & \quad Adaptive & \textbf{47.23}  \hlg{{\scalebox{0.7}{($\uparrow$ +3.2)}}}  & \textbf{59.05}  \hlg{{\scalebox{0.7}{($\uparrow$ +0.8)}}} & \textbf{91.11}  \hlg{{\scalebox{0.7}{($\uparrow$ +0.6)}}} & \textbf{58.60}  \hlg{{\scalebox{0.7}{($\uparrow$ +1.8)}}} \\
& \quad All & 45.26 \scalebox{0.7}{($\uparrow$ +1.2)}& 58.00 \scalebox{0.7}{($\downarrow$ -0.1)} & 88.89 \scalebox{0.7}{($\downarrow$ -0.6)} & 56.70 \scalebox{0.7}{($\downarrow$ -0.1)} \\
\hline
\hline
 & CoT & 16.38 & 30.37 & 58.89 & 22.60 \\
 & L2M & 18.73 & 22.13 & 53.89 & 17.90 \\
 & PoT & 23.73 & 43.77 & 61.11 & 34.50 \\
SmolLM 1.7B & Combined & \underline{24.56} & \underline{46.77} & \underline{67.22} & \underline{35.90} \\
 & \ours\ & & & & \\
  & \quad Adaptive & \textbf{27.98} \hlg{{\scalebox{0.7}{($\uparrow$ +3.4)}}} & 47.20 \scalebox{0.7}{($\uparrow$ +0.4)} & \textbf{72.22} \hlg{{\scalebox{0.7}{($\uparrow$ +5.0)}}} & 37.80 \scalebox{0.7}{($\uparrow$ +1.9)} \\
 & \quad All & 27.75 \scalebox{0.7}{($\uparrow$ +3.2)} & \textbf{48.16} \hlg{{\scalebox{0.7}{($\uparrow$ +1.4)}}} & \textbf{72.22} \hlg{{\scalebox{0.7}{($\uparrow$ +5.0)}}} & \textbf{39.50} \hlg{{\scalebox{0.7}{($\uparrow$ +3.6)}}} \\
\hline
\hline
 & CoT & 55.57 & 68.76 & \underline{99.44} & 66.30 \\
 & L2M & 54.59 & 63.69 & 96.67 & 62.30 \\
 & PoT & 64.22 & 66.94 & 95.56 & \underline{74.30} \\
Qwen 1.5B & Combined & \underline{64.44} & \underline{67.64} & 98.89 & 73.20 \\
 & \ours\ & & & & \\
 & \quad Adaptive & \textbf{64.97} \hlg{{\scalebox{0.7}{($\uparrow$ +0.5)}}} & \textbf{68.98} \hlg{{\scalebox{0.7}{($\uparrow$ +1.3)}}} & \textbf{99.44} \scalebox{0.7}{(-)} & \textbf{75.40} \hlg{{\scalebox{0.7}{($\uparrow$ +1.1)}}} \\
 & \quad All & 64.14 \scalebox{0.7}{($\downarrow$ -0.3 )} & 68.55 \scalebox{0.7}{($\uparrow$ +0.9)} &  98.89 \scalebox{0.7}{($\downarrow$ -0.5)} & 73.50 \scalebox{0.7}{($\downarrow$ -0.8)} \\
\hline
\hline
 & CoT & 36.47 & 54.66 & 83.89 & 43.00 \\
 & L2M & 33.59 & 49.76 & 76.67 & 44.60 \\
 & PoT & 41.62 & 56.83 & \underline{92.22} & \underline{51.40} \\
Qwen 0.5B & Combined & \underline{42.38} & \underline{57.79} & 90.56 & \underline{51.40} \\
 & \ours\ & & & & \\
 & \quad Adaptive & \textbf{43.14} \hlg{{\scalebox{0.7}{($\uparrow$ +0.8)}}}  & 58.44 \scalebox{0.7}{($\uparrow$ +0.7)}  & \textbf{93.33} \hlg{{\scalebox{0.7}{($\uparrow$ +1.1)}}} & 51.70 \scalebox{0.7}{($\uparrow$ +0.3)} \\
 & \quad All & 43.06 \scalebox{0.7}{($\uparrow$ +0.8)}& \textbf{58.61} \hlg{{\scalebox{0.7}{($\uparrow$ +0.8)}}} & 92.22 \scalebox{0.7}{(-)} & \textbf{52.10} \hlg{{\scalebox{0.7}{($\uparrow$ +0.7)}}} \\
\hline
\end{tabular}
}
\label{tab:main-table}
\end{table*}

\paragraph{LLM Based Distillation}
We start by distilling smaller models using the reasoning dataset generated using the LLM in two variations: using data from a single strategy (CoT, PoT, or L2M), and a combination of all three strategies (referred to as ``Combined'').
\autoref{tab:main-table} compares the accuracies of the approaches across four mathematical datasets. The ``Combined'' approach benefited smaller models, yielding slight improvements for the Qwen 0.5B, Qwen 1.5B, and SmolLM 1.7B models. However, it showed little to no improvement, and sometimes even worse performance, for the larger Gemma 2B and 7B models. This indicates that simply merging the distillation data for each strategy is not sufficient for effective multi-strategy distillation.

\paragraph{Consistent improvement across in-distribution dataset}
Compared to the traditional LLM-based distillation approaches, we observe consistent improvements with \ours\ across all models, ranging from 0.5B to 7B parameters as \hlg{highlighted} in \autoref{tab:main-table}. On the in-distribution GSM8K dataset, both Gemma 2B and 7B show significant gains of +3.2 points and +2.5 points respectively (44.05 $\rightarrow$ 47.23 and 71.34 $\rightarrow$ 73.84, respectively). Similarly, SmolLM showed the largest improvement of +3.4 points (24.56 $\rightarrow$ 27.98). Similarly, the smaller Qwen models see gains of +0.5 points for the larger variant (1.5B) and +0.8 points for the smaller variant (0.5B). 

\paragraph{\ours\ performs well on out-of-distribution datasets}
For the out-of-distribution datasets, there is a steady improvement on the ASDiv dataset, with Gemma 7B gaining +1.6 points (69.11 $\rightarrow$ 70.76), +0.8 points for Gemma 2B (58.13 $\rightarrow$ 59.05), +1.4 points for SmolLM (46.77 $\rightarrow$ 48.16), +1.3 points for Qwen 1.5B (67.64 $\rightarrow$ 68.98), and +0.8 points for Qwen 0.5B  (57.79 $\rightarrow$ 58.61). A similar trend is seen for the MultiArith dataset, where SmolLM shows the largest gain of +5 points. It is followed by Qwen 0.5B with +1.1 points, while other models outperform the baseline. In particular, Gemma 7B achieves a perfect score of 100. The results are similar for the SVAMP dataset, with Qwen 0.5B, Qwen 1.5B, SmolLM 1.7B, Gemma 2B, and Gemma 7B gaining +0.7, +1.1, +3.6, +1.8 and +1.4 points, respectively. 

\paragraph{Biasing \ours\ in favor of our strategy of choice}
\begin{figure}
    \centering
    \includegraphics[width=1.0\linewidth]{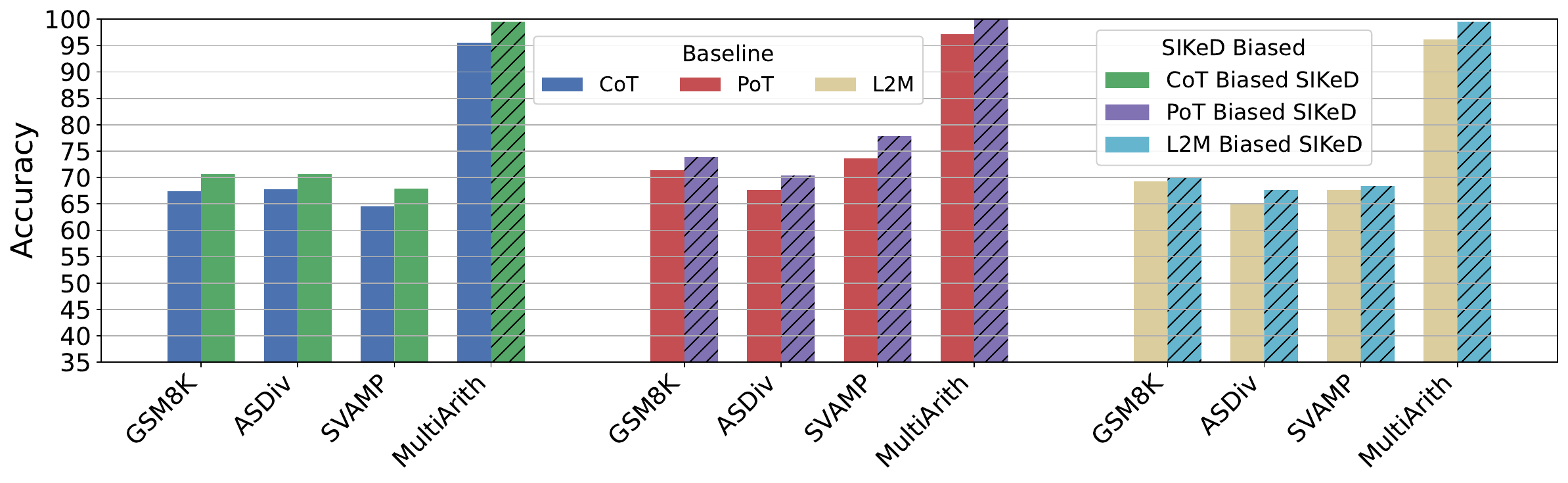}
    \caption{Accuracy comparison between single distillation strategies of CoT, PoT, and L2M with \ours\ biased training using the same strategy using the Gemma 7B model.}
    \label{fig:biasvsSIKeD}
\end{figure}

For some tasks, one strategy might be better than the others due to its performance, lower cost, or better suitability for some use cases (for example, PoT is significantly better for SVAMP compared to other strategies). In such cases, it would be beneficial to bias the selection towards that strategy \footnote{Note that this is different from the already biased selection of the smaller model, as our biased strategy may not be the default biased choice of the smaller model.}. This can be done by taking only the sample from our choice of biased strategy when more than one strategy is correct from the model-generated samples. For example, if for a given data point, a smaller model samples both CoT and PoT correctly, and our biased strategy choice is PoT, we will ignore the CoT output and take only the PoT. \autoref{fig:biasvsSIKeD} compares the individual distillation strategy with the biased \ours. Using Gemma 7B as a smaller model across all datasets, \ours\ outperforms individual distillation strategies by a margin of 2-4 points, highlighting the effectiveness of \ours\ over other distillation approaches.

\paragraph{How many iterations to run for \ours}
\begin{figure}
    \centering
    \includegraphics[width=1.0\linewidth]{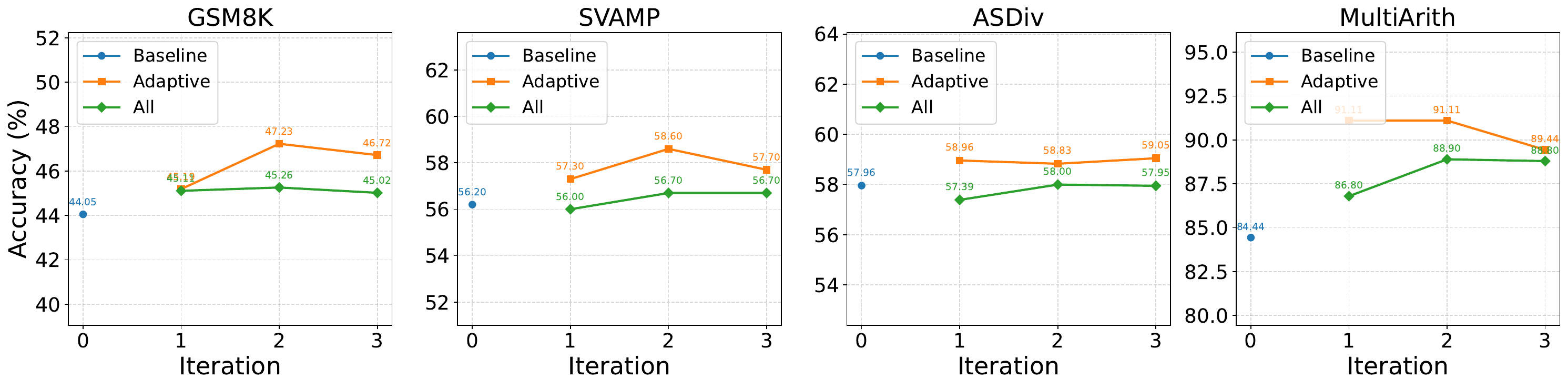}
    \caption{Iterative accuracy comparison for the Gemma 2B model across all datasets. The process is stopped when the gains diminish or when it is no longer cost effective to continue.}
    \label{fig:iterative}
\end{figure}

With each iteration of \ours, the model learns to solve a task using different strategies and adjusts its strategy choice for a given task. This allows for continuous training of \ours. 
\autoref{fig:iterative} illustrates the accuracy improvements across iterations for the Gemma 2B model on various datasets. The iterative training is stopped when accuracy shows only marginal improvements or declines. 
In our experiments, three iterations have consistently proven to be the optimal balance across different models and datasets.

\begin{figure}
    \centering
    \includegraphics[width=1.0\linewidth]{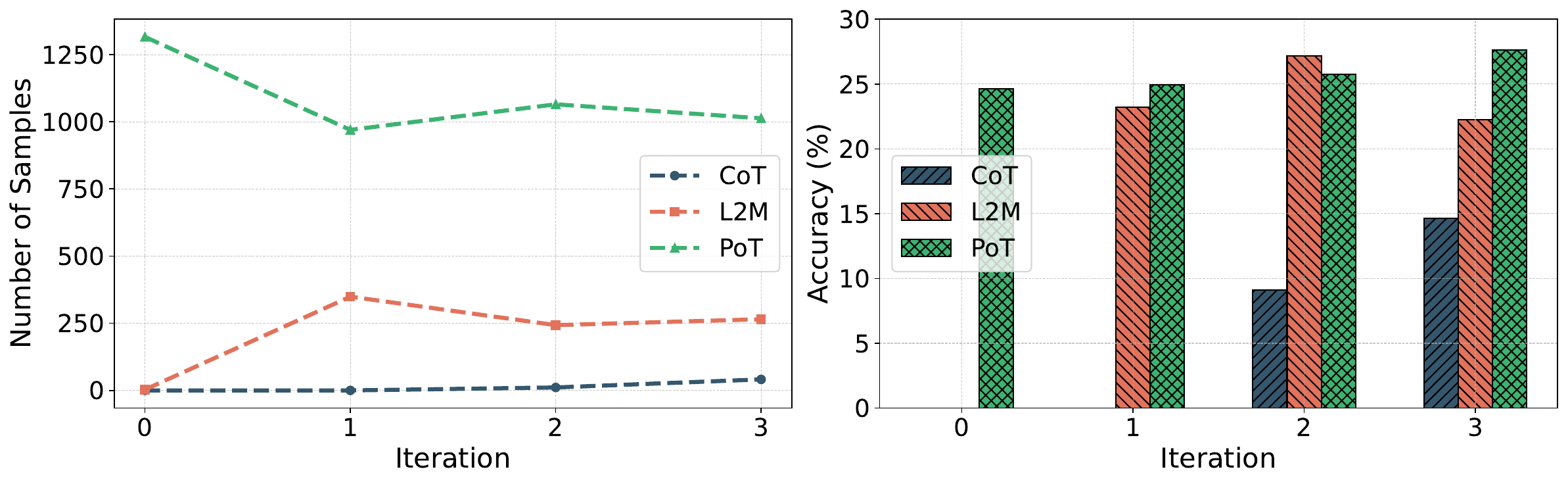}
    \caption{Strategy distribution over iterations for GSM8K dataset using SmolLM 1.7B model.}
    \label{fig:strat-dist}
\end{figure}

\paragraph{How the strategy distribution changes over iterations}
\autoref{fig:strat-dist} illustrates the strategy distribution across different iterations for the GSM8K dataset using the SmolLM 1.7B model. Iteration 0 represents the baseline ``combined'' training from \autoref{tab:main-table}, and as expected, the smaller model is initially biased towards one strategy (PoT in this case). Iterations 1, 2, and 3 show the model’s progression using \ours, where it learns to diversify and select the suitable strategy for the given problem. Notably, while PoT remains the dominant strategy, the model improves its usage of the other two strategies—CoT and L2M—which were absent in the biased baseline. This diversification of strategies results in an overall gain of +3 points over the baseline.

\paragraph{Training from the last checkpoint vs training from pre-trained checkpoint}
In our work, we iteratively train from the last checkpoint using on-policy training as we expect continuous improvements in the model performance with a newly learned strategy. However, an alternative approach uses off-policy training (training the pre-trained model at each iteration) to achieve strong performance \cite{gulcehre2023reinforced}.
We compared on-policy training (our proposed approach) with off-policy training (as in \citet{gulcehre2023reinforced}), utilizing both LLM-generated and self-generated data, and observed a notable decrease in the overall accuracy with off-policy training. Note that we used all of the LLM data at each iteration for off-policy training as the training is done on the pre-trained model. On the GSM8K dataset, our on-policy approach outperformed off-policy training by +6 points (45.26 vs 38.90) using the Gemma 2B model. A similar trend was observed on out-of-distribution datasets, where \ours\ outperformed with a gain of +4-7 points on both the SVAMP and ASDiv datasets (58.6 vs 51.3 for SVAMP and 59.05 vs 55.44 for ASDiv) and a gain of +2 points on the MultiArith dataset (91.11 vs 88.33).

\paragraph{Is data mixing better than LLM-based distillation or self-distillation}

\begin{wrapfigure}{r}{0.55\textwidth}  
    \centering
    \vspace{-0.5cm}
    \includegraphics[width=0.55\textwidth]{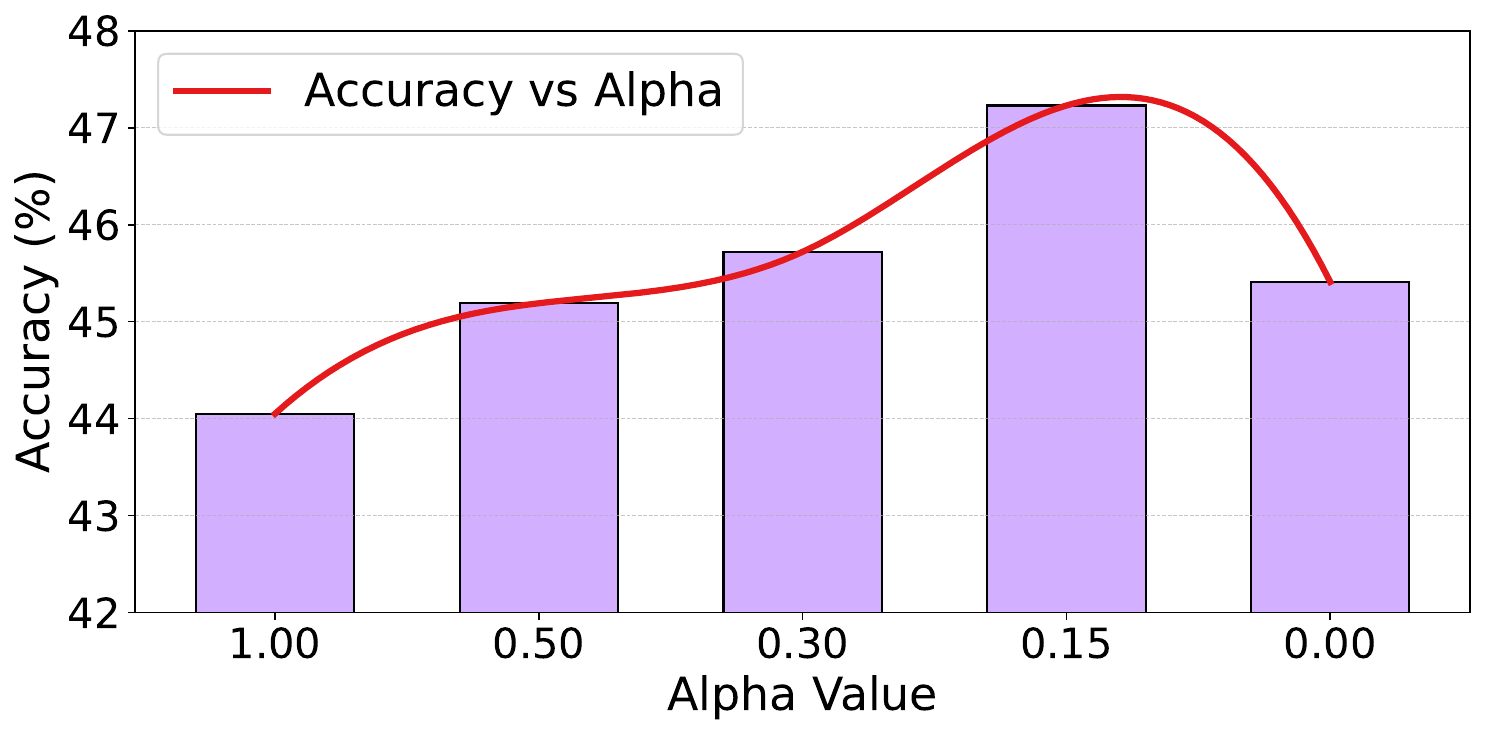}
    \caption{Accuracy comparison with different data mix controlled by $\alpha$. $\alpha=1$ means all data is from the LLM only, while $\alpha=0$ means only smaller model data is used.}
    \label{fig:acc-vs-alpha}
    \vspace{-0.7cm}
\end{wrapfigure}

Mixing data from the smaller model with the LLM helps in bridging the distributional gap between the LLM and the smaller model.
We explore the role of the data mixing rate alpha ($\alpha$) to validate our hypothesis regarding data mixing. When $\alpha$ is set to 1, only LLM-generated data is used, while at the other extreme, $\alpha = 0$ means only data generated by the smaller model is used. As shown in \autoref{fig:acc-vs-alpha}, an $\alpha$ value between 0 and 1 performs better than either of the extreme cases.

\paragraph{Qualitative analysis demonstrating that smaller models learn to choose the right strategy over iterations}
The increase in the performance of smaller models can be attributed to the change in the distribution of reasoning strategies as the model becomes more aligned with its strategy choices. \autoref{fig:qual-strat-change} shows that a model can solve a given problem in the third iteration by switching the reasoning strategy which it initially struggled with. Furthermore, \autoref{fig:qual-improve-reasoning} shows an example of a model correcting its reasoning chain over time due to its improved reasoning capabilities.  

\begin{figure*}[t!]
\fontsize{8}{8.5}\selectfont
\centering
\begin{tabular}{l l l}
\toprule
\addlinespace
\multicolumn{3}{l}{
\parbox{\textwidth}{\textbf{Question} 
There are twice as many boys as girls at Dr. Wertz's school. If there are 60 girls and 5 students to every teacher, how many teachers are there?}} \\
\parbox{0.3\textwidth}{\textbf{Baseline} \\ girls = $60$ \\ 
boys = $2$ $*$ girls \\ 
teachers = girls$/5$ \\ 
answer = teachers \\ 
Final Answer: $12$ 
\textcolor{red}{\ding{55}}} &
\parbox{0.3\textwidth}{\textbf{Iteration1} \\ girls = $60$ \\ 
boys = $2$ $*$ girls \\ 
teachers = girls$/5$ \\ 
answer = teachers \\ 
Final Answer: $12$ 
\textcolor{red}{\ding{55}}} &
\parbox{0.3\textwidth}{\textbf{Iteration2} \\ girls = $60$ \\  
boys = $2$ $*$ girls \\ 
total\_students = girls $+$ boys \\ 
teachers\_per\_student = $5$ \\ 
total\_teachers = total\_students * teachers\_per\_student \\ 
answer = total\_teachers \\ 
Final Answer: $900$ 
\textcolor{red}{\ding{55}}} \\
\addlinespace
\multicolumn{3}{l}{
\parbox{\textwidth}{\textbf{Iteration3} \\ Sub-question 1: How many boys are there? \\ 
Answer to Sub-question 1: $2*60 = <<2*60=120>> 120$ \\
Sub-question 2: How many students are there in total? \\
Answer to Sub-question 2: $60 + 120 = <<60+120=180>> 180$ \\
Sub-question 3: How many teachers are there? \\
Answer to Sub-question 3: $180/5 = <<180/5=36>> 36$ \\
Final Answer: 36 
\textcolor{green}{\ding{51}}}} \\
\addlinespace
\bottomrule
\end{tabular}
\caption{A qualitative example showing how Gemma 2B shifts its strategy selection over iterations to solve a given task.}
\label{fig:qual-strat-change}
\end{figure*}

\begin{figure*}[t!]
\fontsize{9}{9.5}\selectfont
\centering
\begin{tabular}{l}
\toprule
\addlinespace
\addlinespace
\parbox{\textwidth}{\textbf{Question} 
In a dance class of 20 students, 20\% enrolled in contemporary dance, 25\% of the remaining enrolled in jazz dance, and the rest enrolled in hip-hop dance. What percentage of the entire students enrolled in hip-hop dance?} \\
\addlinespace
\parbox{\textwidth}{\textbf{Baseline} \\ 
total\_students = $20$ \\ 
percentage\_enrolled\_contemporary = $0.2$ \\ 
students\_enrolled\_contemporary = total\_students $*$ percentage\_enrolled\_contemporary \\ 
remaining\_students = total\_students $-$ students\_enrolled\_contemporary \\ 
percentage\_enrolled\_jazz = $0.25$ \\ 
students\_enrolled\_jazz = remaining\_students $*$ percentage\_enrolled\_jazz \\ 
remaining\_students = remaining\_students - students\_enrolled\_jazz \\
percentage\_enrolled\_hip\_hop = 1 $-$ percentage\_enrolled\_contemporary $-$ percentage\_enrolled\_jazz \\ 
answer = percentage\_enrolled\_hip\_hop \\ 
Final Answer: $0.55$ 
\textcolor{red}{\ding{55}}} \\
\addlinespace
\parbox{\textwidth}{\textbf{Iteration1} \\ 
total\_students = $20$ \\ 
contemporary\_students = total\_students $*$ $0.2$ \\ 
remaining\_students = total\_students $-$ contemporary\_students \\
jazz\_students = remaining\_students $*$ $0.25$ \\ 
hip\_hop\_students = remaining\_students $-$ jazz\_students \\ 
percentage\_hip\_hop = hip\_hop\_students$/$total\_students $*$ $100$ \\ 
answer = percentage\_hip\_hop \\ 
Final Answer: $60$ 
\textcolor{green}{\ding{51}}} \\
\addlinespace
\addlinespace
\bottomrule
\end{tabular}
\caption{Over iteration, Gemma 2B model learnt to fix its mistake using the same strategy}
\label{fig:qual-improve-reasoning}
\end{figure*}
\section{Related Work}
\paragraph{Knowledge Distillation for Reasoning Tasks} Knowledge distillation \citep{bucilua_model_2006, hinton2015distilling} is a widely-used technique for transferring knowledge from a large language model (LLM) to a smaller model. Previous research has focused on transferring intermediate reasoning steps to smaller models, either step-by-step \citep{shridhar-etal-2023-distilling, magister-etal-2023-teaching, hsieh-etal-2023-distilling} or by combining multiple strategies simultaneously \cite{chenglin2023mixed, zhu2024improvingmath}. These approaches can be viewed as aggregating diverse data sources for distillation, similar to the LLM data approach in our work. In contrast, \cite{hahn_self-knowledge_2019} and \cite{xu_improving_2020} focus on \emph{self-distillation}, where a model improves its performance without external data or knowledge. Specifically, \cite{hahn_self-knowledge_2019} leverages word embeddings, while \cite{xu_improving_2020} uses temporal model checkpoints as a proxy for ground truth. However, both approaches rely solely on data generated by the smaller model and exclude LLM data. Our method strikes a balance between these two extremes by using LLM data to learn multiple strategies and self-generated data to optimize for the right strategy choice.

\paragraph{Self-learning} Previous studies, such as \citep{he_revisiting_2019, sun_self-training_2020, gulcehre2023reinforced, smart}, have shown the effectiveness of the \emph{self-training paradigm} in NLP tasks but are limited by the choice of tasks. While ReST \cite{gulcehre2023reinforced} uses off-policy training, we find on-policy training more suitable for our case both in terms of data efficiency and performance. On-policy training also allows a better choice of learning strategies, since the model can use its most recent learning.  \citet{agarwal2024onpolicy} introduces Generalized Knowledge Distillation (GKD), an on-policy training method that aligns the distributions of large language models (LLMs) and smaller models by incorporating output sequences sampled from the student during training. However, the task was limited to the distribution alignment and not to aligning the strategy choices in a multi-strategy distillation. Simply applying GKD would not address this issue, as it would force the smaller model to learn all strategies, which is impractical given its limited capacity. 

Finally, we compare our distillation strategies with LLM-based distillation using both individual strategies \cite{shridhar-etal-2023-distilling, magister-etal-2023-teaching, hsieh-etal-2023-distilling} and a combination of several strategies at once \cite{chenglin2023mixed, zhu2024improvingmath}.
\section{Conclusion}
We propose \oursfull, that addresses the challenge of distilling multistep reasoning skills from large language models (LLMs) to smaller models. Unlike traditional methods, which often leave smaller models biased towards a single strategy, \ours\ uses iterative \emph{self-guided} training, combining LLM and self-generated data to improve overall reasoning in smaller models. We demonstrate our approach across various mathematical reasoning datasets and demonstrate that \ours\ improves the ability of smaller models to handle complex reasoning, achieving significant performance gains.

\bibliography{main}

\newpage

\section{Appendix}

\label{app:prompts}
\begin{figure*}[h]
\small
\centering
\begin{tabular}{l}
\toprule
\addlinespace
\parbox{\textwidth}{Below is an instruction that describes a task, paired with an input that provides further context. Write a response that appropriately completes the request.
\newline
\newline
Instruction:
Solve the given math problem step by step. Put your final answer after 'Final answer:'.
\newline
\newline
Input:
John buys 3 dress shirts.  They sell for \$20 each. He also has to pay 10\% tax on everything. How much did he pay in total?
\newline
\newline
Response:
The shirts cost 3*\$20=\$$<<$3*20=60$>>$60 before tax
The tax cost \$60*.1=\$$<<$60*.1=6$>>$6
So in total they paid \$60+\$6=\$$<<$60+6=6$>>$ 66
Final Answer: 66$<$eos$>$} \\
\addlinespace
\addlinespace
\parbox{\textwidth}{\centering \bf  [7 more examples randomly sampled from the training set]}\\
\addlinespace
\addlinespace
\parbox{\textwidth}{
Input:
Thomas is training at the gym to prepare for a competition. He trained for 5 hours every day for a month (30 days). If he continues to train for the next 12 days, how many hours will he spend on training in total?
\newline
\newline
Response:
\textcolor{teal}{Model generated response ..}} \\
\bottomrule
\end{tabular}
\caption{8-shot Chain of Thought demonstration.}
\label{fig:prompt-8shot-cot}
\end{figure*}

\begin{figure*}[h!]
\small
\centering
\begin{tabular}{l}
\toprule
\addlinespace
\parbox{\textwidth}{Below is an instruction that describes a task, paired with an input that provides further context. Write a response that appropriately completes the request.
\newline
\newline
Instruction:
Solve the given math problem by decomposing it into smaller, manageable sub-questions. Put your final answer after 'Final answer: '.
\newline
\newline
Input:
John buys 3 dress shirts.  They sell for \$20 each. He also has to pay 10\% tax on everything. How much did he pay in total?
\newline
\newline
Response:
Sub-question 1: How much did John spend on shirts? \\
Answer to Sub-question 1: 3*\$20=\$$<<$3*20=60$>>$60 \\
Sub-question 2: How much did John spend on tax? \\
Answer to Sub-question 2: \$60*.1=\$$<<$60*.1=6$>>$6 \\
Sub-question 3: How much did John spend in total? \\
Answer to Sub-question 3: \$60+\$6=\$$<<$60+6=66$>>$66 \\
Final Answer: 66$<$eos$>$} \\
\addlinespace
\addlinespace
\parbox{\textwidth}{\centering \bf  [7 more examples randomly sampled from the training set]}\\
\addlinespace
\addlinespace
\parbox{\textwidth}{
Input:
Thomas is training at the gym to prepare for a competition. He trained for 5 hours every day for a month (30 days). If he continues to train for the next 12 days, how many hours will he spend on training in total?
\newline
\newline
Response:
\textcolor{teal}{Model generated response ..}} \\
\bottomrule
\end{tabular}
\caption{8-shot Least to Most demonstration.}
\label{fig:prompt-8shot-l2m}
\end{figure*}

\begin{figure*}[h!]
\small
\centering
\begin{tabular}{l}
\toprule
\addlinespace
\parbox{\textwidth}{Below is an instruction that describes a task, paired with an input that provides further context. Write a response that appropriately completes the request.
\newline
\newline
Instruction:
Solve the given math problem by writing a python program. Store your result as a variable named 'answer'.
\newline
\newline
Input:
John buys 3 dress shirts.  They sell for \$20 each. He also has to pay 10\% tax on everything. How much did he pay in total?
\newline
\newline
Response:
total\_shirts = 3 \\
cost\_of\_one\_shirt = 20 \\
total\_cost\_shirts = total\_shirts * cost\_of\_one\_shirt \\
tax\_rate = 0.1  \\
tax\_amount = tax\_rate * total\_cost\_shirts \\
total\_cost = total\_cost\_shirts + tax\_amount \\
answer = total\_cost $<$eos$>$} \\
\addlinespace
\addlinespace
\parbox{\textwidth}{\centering \bf  [7 more examples randomly sampled from the training set]}\\
\addlinespace
\addlinespace
\parbox{\textwidth}{
Input:
Thomas is training at the gym to prepare for a competition. He trained for 5 hours every day for a month (30 days). If he continues to train for the next 12 days, how many hours will he spend on training in total?
\newline
\newline
Response:
\textcolor{teal}{Model generated response ..}} \\
\bottomrule
\end{tabular}
\caption{8-shot Program of Thought demonstration.}
\label{fig:prompt-8shot-pot}
\end{figure*}

\end{document}